\documentclass[letterpaper, 10 pt, conference]{ieeeconf}  % Comment this line out if you need a4paper

\IEEEoverridecommandlockouts                              % This command is only needed if 
\usepackage{xcolor}
\usepackage{multirow}
\usepackage{gensymb}
\usepackage{booktabs} 
\usepackage{cite}
\usepackage{algorithm}
\usepackage{algorithmic}
\usepackage{makecell}
\usepackage{textcomp}
\usepackage{graphics} % for pdf, bitmapped graphics files
\usepackage{epsfig} % for postscript graphics files
\usepackage{times} % assumes new font selection scheme installed
\usepackage{amsmath} % assumes amsmath package installed
\usepackage{amssymb}  % assumes amsmath package installed
\usepackage{xspace}
\usepackage{graphicx}
\usepackage{amsmath}
\usepackage{amssymb}
\usepackage{booktabs}
\usepackage{tabularx}
\usepackage{multirow}
\usepackage{hyperref}
\usepackage{adjustbox}
\usepackage{tabularray}
\usepackage{dsfont}
\usepackage{rotating}
\usepackage{tikz}
\usepackage{pdflscape}
\usepackage{wrapfig}
\usepackage{lscape}
\usepackage{soul}
\usepackage{subcaption}
\usepackage{caption}

\usepackage{hyperref}

\definecolor{radarpts}{RGB}{253, 124, 22}
\definecolor{lidarpts}{RGB}{56, 182, 58}
\definecolor{spatial_shift}{RGB}{230, 230, 20}
\definecolor{hallucination}{RGB}{255,119,197}
\definecolor{det}{RGB}{56, 182, 58}
\definecolor{select_match}{RGB}{184, 82, 255}
\definecolor{shared_head}{RGB}{0, 230, 220}
\definecolor{det_head}{RGB}{90, 200, 255}
\definecolor{gt}{RGB}{232, 67, 63}
\definecolor{wrong_det}{RGB}{255, 240, 127}
\definecolor{step_1}{RGB}{239, 190, 3}
\definecolor{training_step_1}{RGB}{84, 173, 21}

\title{\LARGE \bf
Robust 3D Object Detection from LiDAR-Radar Point Clouds \\ via Cross-Modal Feature Augmentation
}

\author{Jianning Deng\authorrefmark{2}, Gabriel Chan\authorrefmark{3}\authorrefmark{7}, Hantao Zhong\authorrefmark{4}\authorrefmark{7}, and Chris Xiaoxuan Lu\authorrefmark{5}\authorrefmark{1}
\thanks{\authorrefmark{1}Corresponding author. Email: xiaoxuan.lu@ucl.ac.uk}
\thanks{\authorrefmark{7}Equal Contribution} 
\thanks{\authorrefmark{2}University of Edinburgh, United Kingdom. \authorrefmark{3}Kodifly Limited, Hong Kong. 
\authorrefmark{4}University of Cambridge, United Kingdom. \authorrefmark{5}University College London, United Kingdom. 
}
}

\begin{document}
\maketitle
\thispagestyle{empty}
\pagestyle{empty}

%%%%%% ABSTRACT
%%%%%%%%%%%%%%%%%%%%%%%%%%%%%%%%%%%%%%%%%%%%%%%%%%%%%%%%%%%%%%%%%%%%%%%%%%%%%%%%
\begin{abstract}
This paper presents a novel framework for robust 3D object detection from point clouds via cross-modal hallucination. Our proposed approach is agnostic to either hallucination direction between LiDAR and 4D radar. We introduce multiple alignments on both spatial and feature levels to achieve simultaneous backbone refinement and hallucination generation. Specifically, spatial alignment is proposed to deal with the geometry discrepancy for better instance matching between LiDAR and radar. The feature alignment step further bridges the intrinsic attribute gap between the sensing modalities and stabilizes the training. The trained object detection models can deal with difficult detection cases better, even though only single-modal data is used as the input during the inference stage. Extensive experiments on the View-of-Delft (VoD) dataset show that our proposed method outperforms the state-of-the-art (SOTA) methods for both radar and LiDAR object detection while maintaining competitive efficiency in runtime.
\end{abstract}

%%%%%% BODY TEXT
%%%%%%%%%%%%%%%%%%%%%%%%%%%%%%%%%%%%%%%%%%%%%%%%%%%%%%%%%%%%%%%%%%%%%%%%%%%%%%%%

\section{Introduction}

Robust recognition and localization of objects in 3D space is a fundamental perception task and an essential capability for intelligent systems. In the context of autonomous driving, accurate 3D object detection is vital for safe motion planning, especially in a complex urban environment. Due to accurate depth measurement in long-range and robustness to illumination conditions, ranging sensors such as LiDAR and radar have attracted increasing attention recently and in turn, make the point clouds from them one of the most commonly used data representations for 3D object detection.

Despite advancements in LiDAR-based \cite{pointpillar, pointrcnn, yang2019std, liu2020tanet, shi2020pv, shi2020points, zhang2022not, yang20203dssd, yin2021center, noh2021hvpr} and radar-based 3D object detection \cite{bansal2020pointillism, vod}, each exhibits inherent drawbacks. 
LiDAR excels at producing dense point clouds but lacks per-point dynamic/velocity information. Conversely, the emerging 4D radars, while prone to sparse data, present valuable semantic information per point, such as the Radar Cross Section (RCS) and Doppler velocity. Unlike LiDAR intensity, RCS is a \emph{distance-independent} measurement uniquely determined by the object material and reflection angle. The Doppler velocity measures the moving speed of a detected point relative to the ego vehicle. Benefiting from the rich semantic information, fairly reasonable 3D object performance can be achieved using 4D radars in complex urban environments \cite{vod}, even if some objects have less than 10 radar points.

\begin{figure}[t!]
    \centering
    \begin{subfigure}[h]{0.9\linewidth}
    \includegraphics[width=\linewidth]{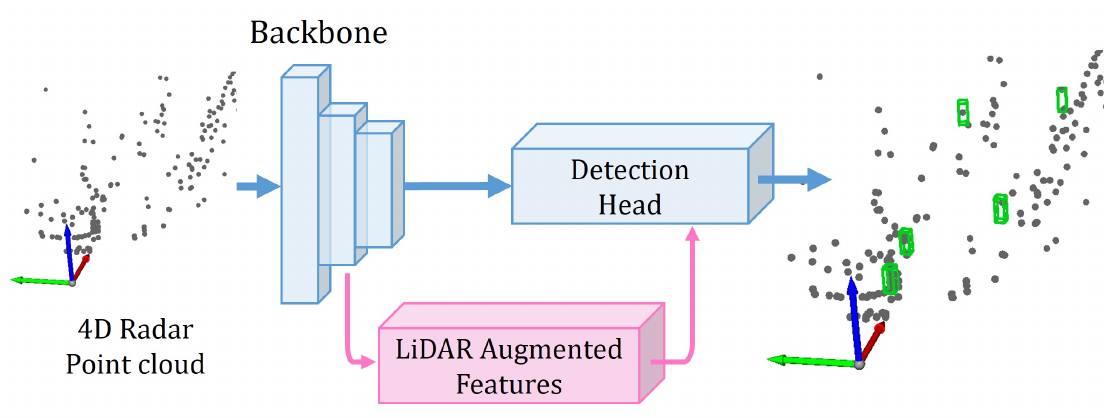}
    \caption{Proposed method for 4D radar object detection}
    \label{fig:proposed_method}
    \end{subfigure}
    \begin{subfigure}[h]{0.45\linewidth}
    \includegraphics[width=\linewidth]{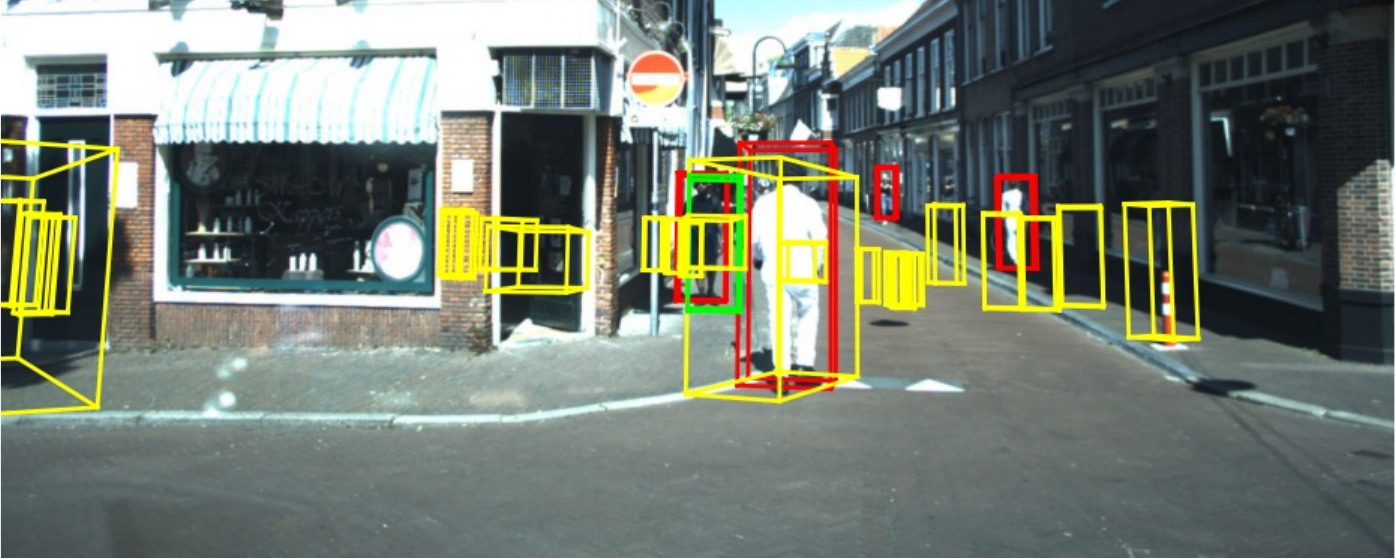}
    \caption{Centerpoint}
    \label{fig:open_fig_ctp}
    \end{subfigure}
    \begin{subfigure}[h]{0.45\linewidth}
    \includegraphics[width=\linewidth]{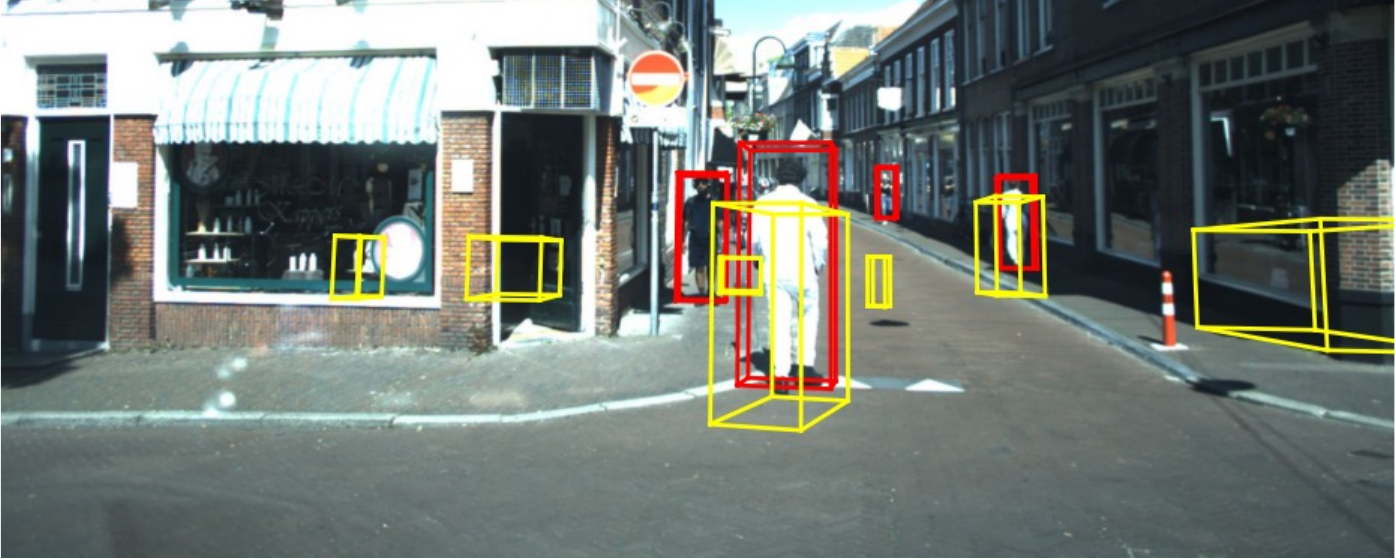}
    \caption{Pointpillar}
    \end{subfigure}
    \begin{subfigure}[h]{0.45\linewidth}
    \includegraphics[width=\linewidth]{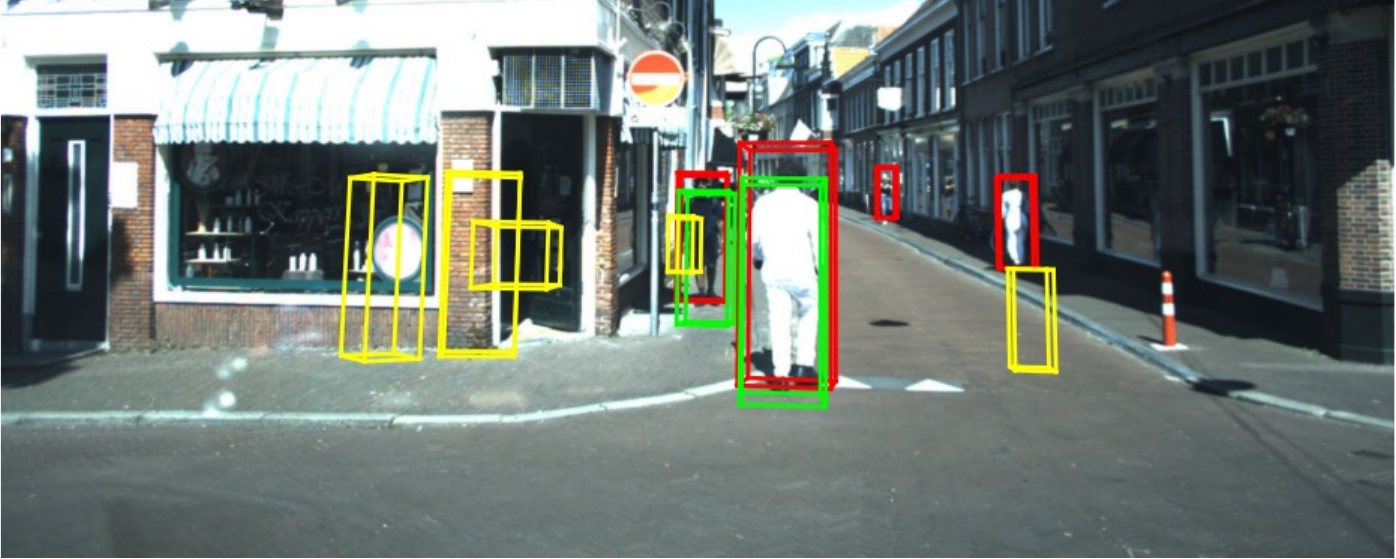}
    \caption{PV-RCNN}
    \end{subfigure}
    \begin{subfigure}[h]{0.45\linewidth}
    \includegraphics[width=\linewidth]{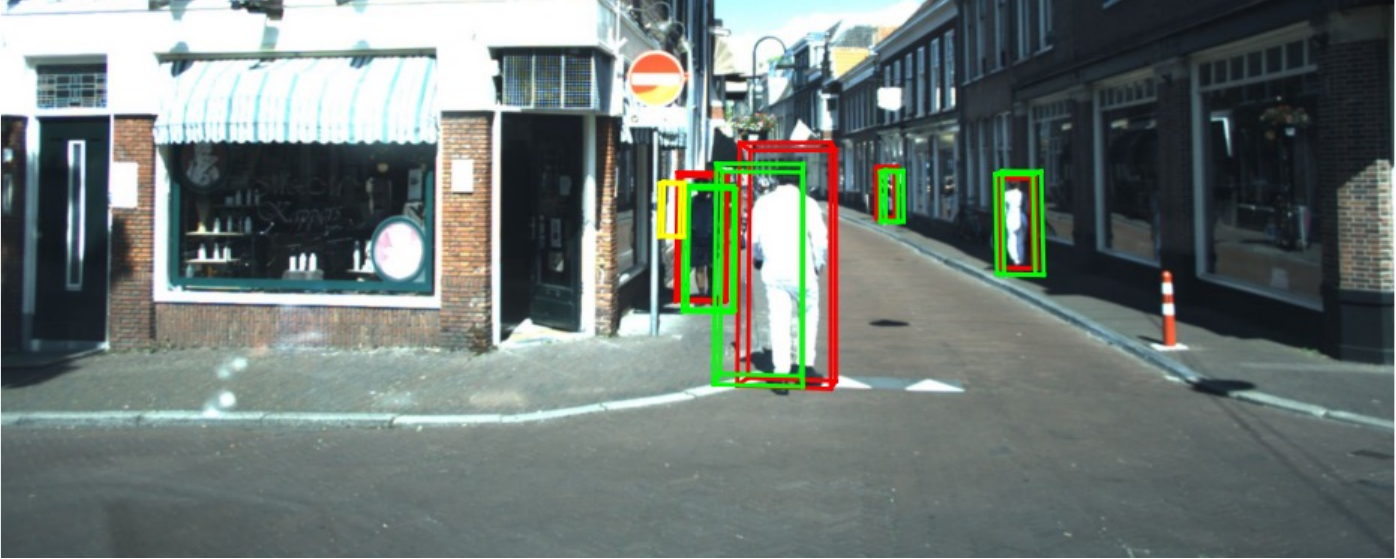}
    \caption{Ours}
    \label{fig:open_fig_ours}
    \end{subfigure}
    \caption{Fig.~\ref{fig:proposed_method} illustrates the proposed method with the 4D radar input as an example. Fig.~\ref{fig:open_fig_ctp} - Fig.~\ref{fig:open_fig_ours} are the visualization of radar detection in the same scene of different methods. Ground truth boxes are denoted in \textcolor{gt}{red}, the false detections are denoted in \textcolor{step_1}{yellow} and the correct detections are denoted in \textcolor{det}{green}. RGB images are \textbf{only used for visualization.}}
    \label{fig:open_fig}
%    \vspace{-5pt}
    \vspace{-2em}
\end{figure}

Given the complementary nature of these sensors, we explore if one can assimilate the characteristics of the other while enhancing independent detection capacities. Indeed, due to the cost consideration, many low-end vehicles are only equipped with either radars or LiDARs. It is thus valuable to train a \emph{single-modal} detection model from the \emph{multimodal} data collected by some pilot vehicles equipped with both sensors, yet dispatch the trained models on low-end vehicles equipped with only one of them.
Addressing this necessitates delving into cross-modal learning. Prior arts achieve this via either feature augmentation \cite{xu2017learning, saputra2020deeptio, hoffman2016learning, lezama2017not, choi2017learning} or knowledge distillation for backbone refinement \cite{zheng2022boosting, yuan2022x, gupta2016cross, ren2021learning}. However, these works are dedicated to images and the learning direction is \emph{unidirectional}, e.g., RGB camera $\rightarrow$ depth camera. 

In this work, we study cross-modal feature augmentation for point cloud inputs and propose a new learning framework that can work in agnostic directions (i.e., LiDAR $\leftrightarrow$ radar). We introduce a novel selective matching module to bridge the intrinsic sensing gap between two modalities to enable feature-level alignment in the shared latent space.
Our design enhances intra-modality features through cross-modal learning and supplements inter-modality features for improved detection robustness (see Fig.~\ref{fig:open_fig}). Our method outperforms previous state-of-the-art (SOTA) methods in 3D object detection, which we demonstrate on the public View-of-Delft dataset in both LiDAR and radar object detection tasks. Our code is publicly released at \href{https://github.com/DJNing/See_beyond_seeing}{\color{blue}{https://github.com/DJNing/See\_beyond\_seeing}}.

\begin{figure*}[t!]
  \centering
  \includegraphics[width=0.95\textwidth]{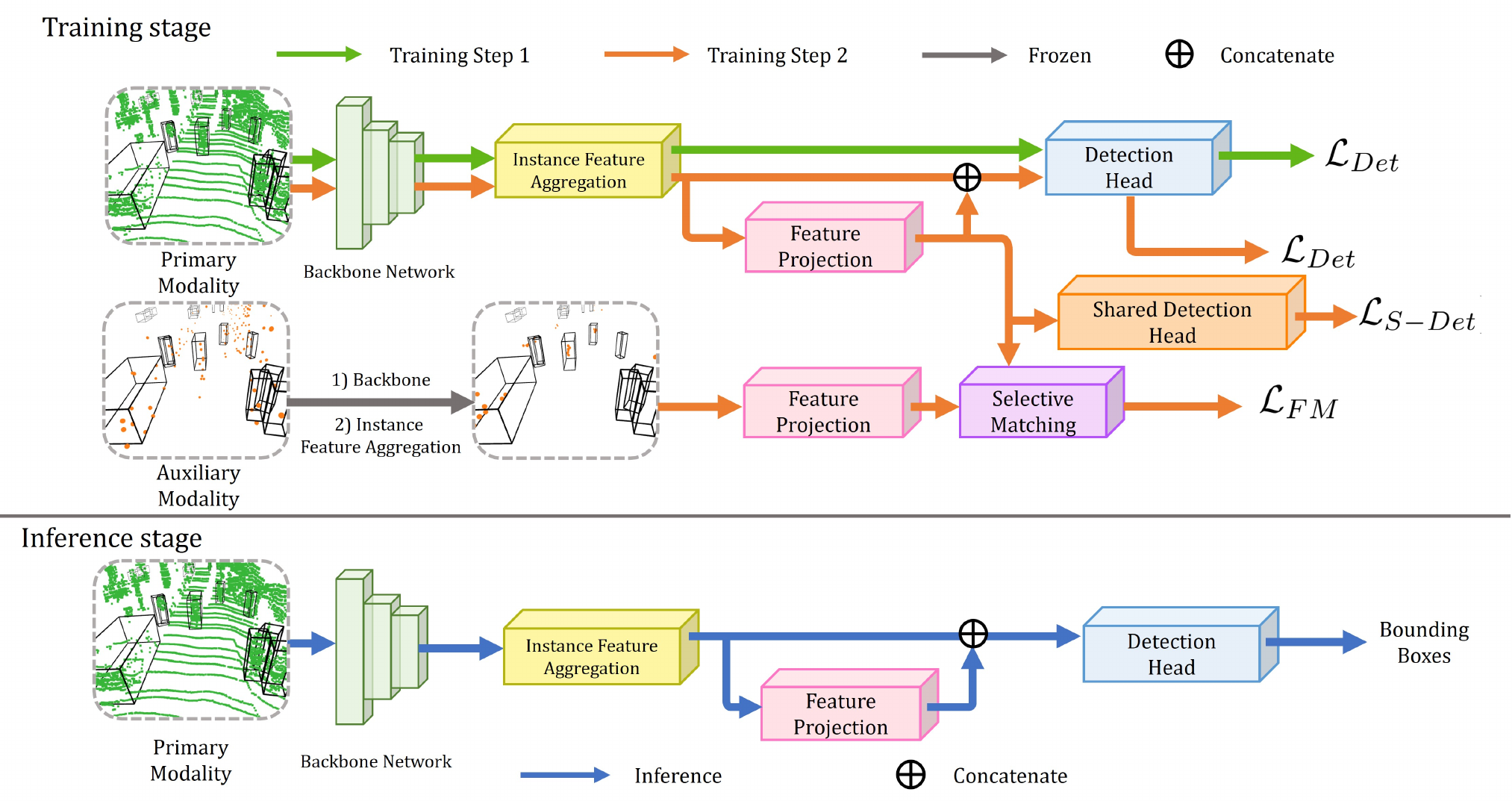}
  \caption{Method Overview. The upper figure illustrates the 2-step training strategy, blocks used in the first step of training are connected with \textcolor{training_step_1}{green line} and those for the second step are connected with \textcolor{orange}{orange line}. Note the primary and auxiliary data can be interchangeable among two sensor modalities (radar and LiDAR) depending on the end goal. \textbf{Only single modal data (primary modal) will be used during inference} as shown in the lower figure connected with \textcolor{blue}{blue line}. }
  \label{fig:overview}
  \vspace{-1.5em}
\end{figure*}
\section{Related work}

\noindent \textbf{LiDAR object detection}. LiDAR 3D object detection is categorized into voxel-based, point-based, and hybrid (i.e., point-voxel) approaches. 
Voxel-based methods transform point cloud data into grid cells, suitable for convolution operations \cite{zhou2018voxelnet, yan2018second,yin2021center,pointpillar}. For example, \cite{pointpillar} converts point clouds into 2D pseudo-images for faster inference, while \cite{yin2021center} adopts a two-stage process, improving detection at a higher memory cost.
Point-based techniques directly process unstructured point cloud data, minimizing data loss \cite{pointrcnn, yang20203dssd, zhang2022not, shi2020point}. Models in this category, such as \cite{pointrcnn, zhang2022not}, utilize deep sets methodologies from \cite{qi2017pointnet} and \cite{qi2017pointnet++} to extract features.
Recent research introduces point-voxel networks that combine the best of both methods for enhanced accuracy \cite{shi2020pv, yang2019std, noh2021hvpr, he2020structure}. As highlighted by \cite{zhang2022not}, these networks, like \cite{shi2020pv}, perform well on the KITTI dataset \cite{kitti}, though they require more computation resources.

% \subsection{}
\noindent \textbf{Radar object detection}. Early efforts in radar-based object detection focused on 2D object detection \cite{wang2021rodnet, zhang2021raddet, qian2021robust}. It is only with the recent availability of 4D radar sensors that radar 3D object detection began receiving attention from researchers. \cite{bansal2020pointillism} is an anchor-based 3D detection framework with a PointNet style backbone. It was only evaluated on short-range radar point clouds within $\sim$ 10 meters in front of the ego-vehicle, which is far too close for real-world autonomous scenarios. On the other hand, authors of the recently published dataset \cite{vod} successfully repurposed PointPillars \cite{pointpillar} on radar point clouds. Although the 3D object detection architecture was designed for LiDAR point clouds, it achieved SOTA performance on their radar 3D object detection benchmark \cite{vod}.

\noindent \textbf{Cross-modal feature augmentation}.  Learning with heterogeneous sensor information involves using supplementary data during training to enhance single-modality networks \cite{saputra2020deeptio, lezama2017not, choi2017learning, xu2017learning, hoffman2016learning, yuan2022x, gupta2016cross, zheng2022boosting}. \cite{hoffman2016learning} pioneered using cross-modal learning in object detection by enhancing RGB image detection using depth image features. \cite{xu2017learning} integrated thermal data for RGB pedestrian detection. Recent works, like \cite{zheng2022boosting}, utilize multi-modal features via knowledge distillation to improve 3D detection. However, many of these methods do not maximize cross-modal information use. For instance, \cite{xu2017learning, hoffman2016learning} require an extra backbone for hallucination instead of optimizing the original. Knowledge distillation techniques \cite{yuan2022x, gupta2016cross, zheng2022boosting}, while strengthening the backbone, are limited to image inputs.
\section{Method}

The proposed method is depicted in Fig.~\ref{fig:overview}. Our framework includes (i) a point-based backbone for feature extraction, (ii) an instance feature aggregation module for aligning input modalities, (iii) a feature projection branch for better cross-modal alignment, and (iv) a detection head for bounding box prediction. The selective matching module and shared detection head are training-specific. We use `primary' and `auxiliary' interchangeably to denote our two sensor modalities for clarity. At a high-level summary of our method, a backbone network takes point set \(\mathbf{P}\) of the primary modality and outputs a subset \(\mathbf{P}_f\) for the sampled foreground points with extracted features. For the auxiliary-modal data, we use a \textbf{hat} notation, such as \(\hat{\mathbf{P}}\). The instance feature aggregation module processes \(\mathbf{P}_f\) to obtain `centered points' $\mathbf{C}$, adjusting point positions with a spatial offset \(\tilde{o_i}\). Features from both modalities are then aligned via non-linear mapping, constrained by instance location, resulting in inter-modality augmented features \(\mathbf{H}\). These are then concatenated with \(\mathbf{C}\) and fed into the detection head for final object recognition.

\subsection{Backbone Network}
We construct our backbone network based on the Set-Abstraction (SA) layer proposed in PointNet++ \cite{qi2017pointnet++} for better efficiency and to avoid information loss \cite{yang20203dssd, zhang2022not}. Additionally, the farthest-point sampling (FPS) operation in SA layer is replaced with center-aware sampling proposed in \cite{zhang2022not} for better performance, which selects top-$k$ points based on the predicted centeredness. This predicted centeredness is constrained during training as:
\begin{equation}
\resizebox{0.9\linewidth}{!}{
    $\mathcal{L}_{Ctr} = -\sum^{N}_{k=1}(Mask_{k} \cdot y_k log(\tilde{y}_k) + (1-y_k)log(1-\tilde{y}_k))$}
     \label{eq:ctr-loss}
\end{equation}
with $y_k$ as the ground truth centeredness and $\tilde{y}_k$ as network estimated value. $Mask_k$ is the corresponding mask value for each point proposed in \cite{yang20203dssd}, which assigns higher weights to points closer to the centroid of objects, and no weight at all for background points. The output of the backbone is denoted as $\mathbf{P}_f = \{p_i\} \in \mathbb{R}^{N_f\times (3+D)}$ with $N_f $ points and $p_i = [t_i, f_i^D]$.

\begin{figure}[t!]
  \centering
  \begin{subfigure}[b]{0.3\linewidth}
    \includegraphics[width=\linewidth]{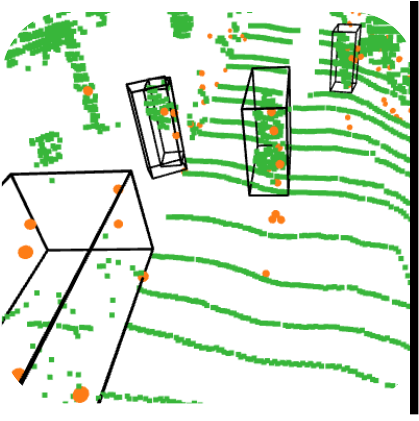}
    \caption{Inputs}
    \label{fig:sparsity_input}
  \end{subfigure}
  \begin{subfigure}[b]{0.32\linewidth}
    \includegraphics[width=\linewidth]{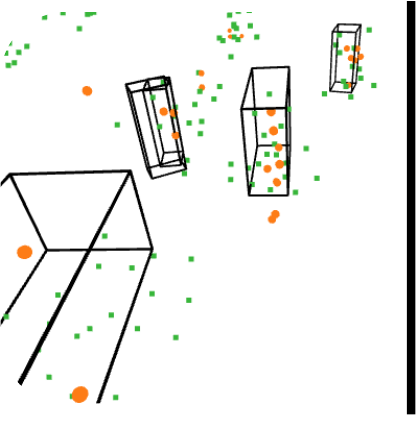}
    \caption{Backbone results}
    \label{fig:sparsity_sampled}
  \end{subfigure}
  \begin{subfigure}[b]{0.3\linewidth}
    \includegraphics[width=\linewidth]{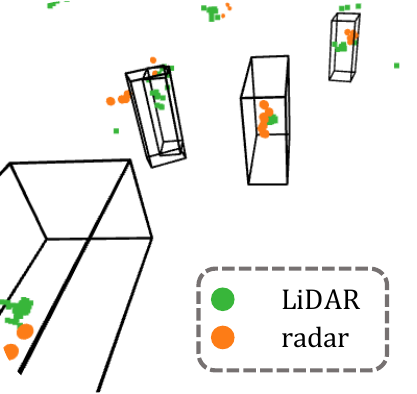}
    \caption{Centered}
    \label{fig:sparsity_centre}
  \end{subfigure}
  \caption{Point-level matches are difficult to obtain without centroid generation in the instance feature aggregation module due to sparsity. The black bounding boxes are the object ground truth labels. (Best viewed in color and zooming in).}
  \label{fig:sparsity}
  \vspace{-1em}
\end{figure}

\subsection{Instance Feature Aggregation}
\label{sec:spatial_alignment}

Single-stage detectors face challenges in point matching across different sensor modalities due to discrepancies between point clouds from co-located LiDAR and radar, evident in Fig.\ref{fig:sparsity_input}. This difference, arising from fundamental sensor characteristics and unsolvable by calibration, is worsened by sampling randomness, especially in the backbone's foreground sampling (c.f. Fig.~\ref{fig:sparsity_sampled}).
To mitigate this, we use instance feature aggregation for improved alignment before cross-modal matching. This process is illustrated in Fig.~\ref{fig:sparsity_centre}. Following \cite{qi2019deep}, we generate `centred points' for context. Each foreground point, $p_i \in \mathbf{P}_f$, gets an offset, $\tilde{o}_i \in \mathbb{R}^3$, guiding it towards its object center. We constrain this regression process with a smooth-L1 loss as:
\begin{equation}
	\mathcal{L}_{O-Reg} = \frac{1}{\sum_i \mathbf{I}(p_i)} \sum_{i=1}^m smooth_{L1}(o_i - \tilde{o}_i) \cdot \mathbf{I}(p_i)
    \label{eqn:vote_loss}
\end{equation}
here $o_i$ is the true offset, and $\mathbf{I}(p_i)$ checks if $p_i$ is within an object. Points after this instance-aggregation step is denoted as $\mathbf{C} = \{c_i\} \in \mathbb{R}^{N_f\times(3+D)}$, with each having a new position $t_i' = t_i + \tilde{o_i}$. Since points now cluster near the instance centroid, the subsequent SA layer's ball query often groups the same points, resulting in almost identical instance features after max pooling. Thus, we can quickly establish instance matches between modalities by matching points in close spatial locations.

%  Thus, we can quickly establish instance matches between modalities by matching points in close spatial locations. 
% These clustered points with shifted coordinates towards their centroids will be fed to the hallucination branch and detection head subsequently. 

% \begin{figure}[t!]
%     \centering
%     \includegraphics[width=0.9\linewidth]{figure/self_cluster.pdf}
%     \caption{\hant{This Fig could be deleted}Illustration of the centroid generation. The left figure is the input point clouds. The middle one shows the sampling process of the backbone network. Foreground points with high confidence (points in \textcolor{lidarpts}{green}) survived in the final layers. The right figure demonstrates the centroid generation process, where foreground points are moved toward the object center.}
%     \label{fig:self_clustering}
%     \vspace{-1em}
% \end{figure}

\subsection{Alignment-aware Feature Projection}
\label{sec:feature_alignment}
Different sensing principles make LiDAR and radar instance features modality-specific. For instance, radar measures relative Doppler velocity \cite{meyer2019deep, meyer2019automotive, caesar2020nuscenes}, while LiDAR provides detailed object geometry. Given these intrinsic differences, it is impractical to directly match all features across modalities. A solution is to ground cross-modal learning in a shared subspace, focusing on a subset of their features (the \textcolor{hallucination}{magenta block} in Fig.~\ref{fig:overview}). This mandates a feature projection module, allowing feature embedding in a shared latent space for both modalities.
Specifically, given the instance-aggregated point set $\mathbf{C} \in \mathbb{R}^{N_f \times (3+D)}$ of the primary modality in the domain $X$ and clustered point set $\hat{\mathbf{C}} \in \mathbb{R}^{\hat{N}_f \times (3 + \hat{D})} $ of the auxiliary modality in domain $\hat{X}$, we have two mapping functions acting as the feature projection: $F_{pri}: X \rightarrow H$ and $F_{aux}: \hat{X} \rightarrow \hat{H}$ to project both modalities to a shared common space. We adopt the MLP block for each of the mapping functions which yields two inter-modality augmented feature sets $\mathbf{H} = \{h_i\} \in \mathbb{R}^{N_f \times F}$ and $\mathbf{\hat{H}} = \{h_i\} \in \mathbb{R}^{\hat{N_f} \times F}$ respectively. The dimension of the shared common subspace, $F$, is set empirically. 

\subsection{Selective Matching}
\begin{figure}[t!]
    \centering
    \begin{subfigure}[b]{\linewidth}
        \centering
        \includegraphics[width=0.9\linewidth]{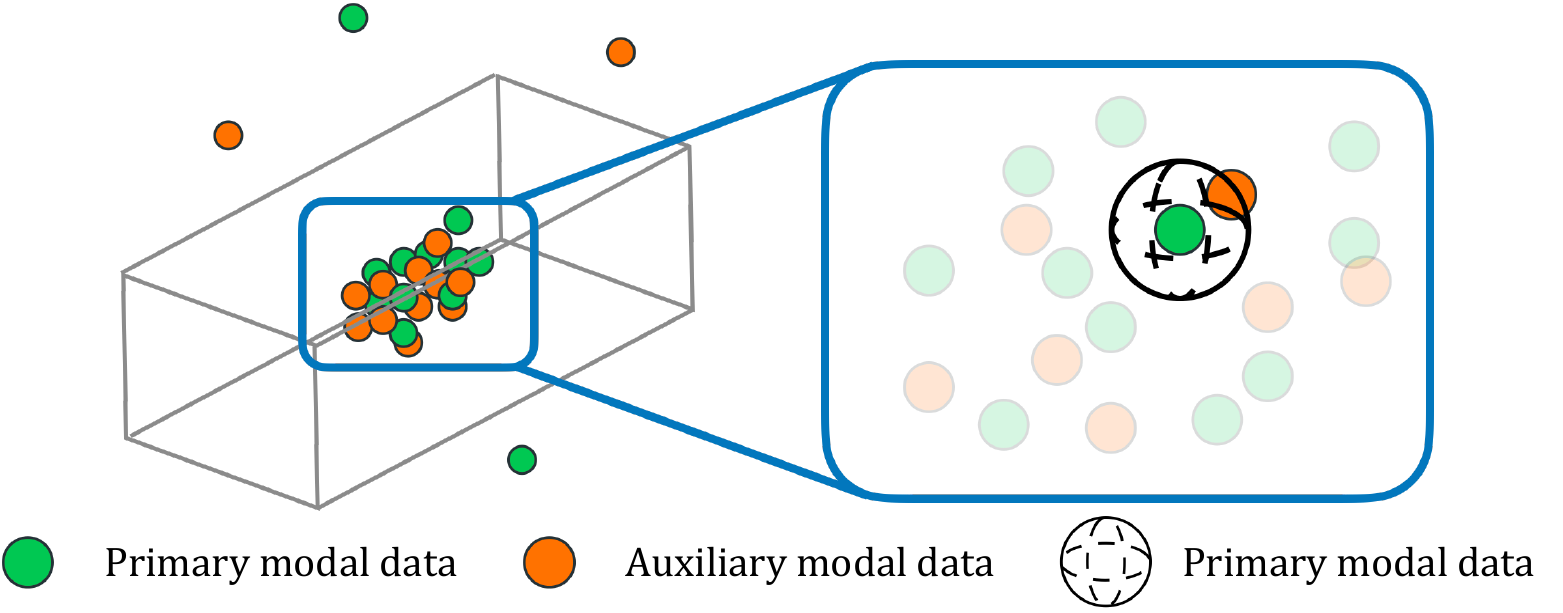}
        \caption{Illustration of selective matching. Points with lighter colors on the right are ignored in the match-searching process. }
        \vspace{0.5em}
        \label{fig:selective_matching}
    \end{subfigure}
    \begin{subfigure}[b]{\linewidth}
        \centering
        \includegraphics[width=0.95\linewidth]{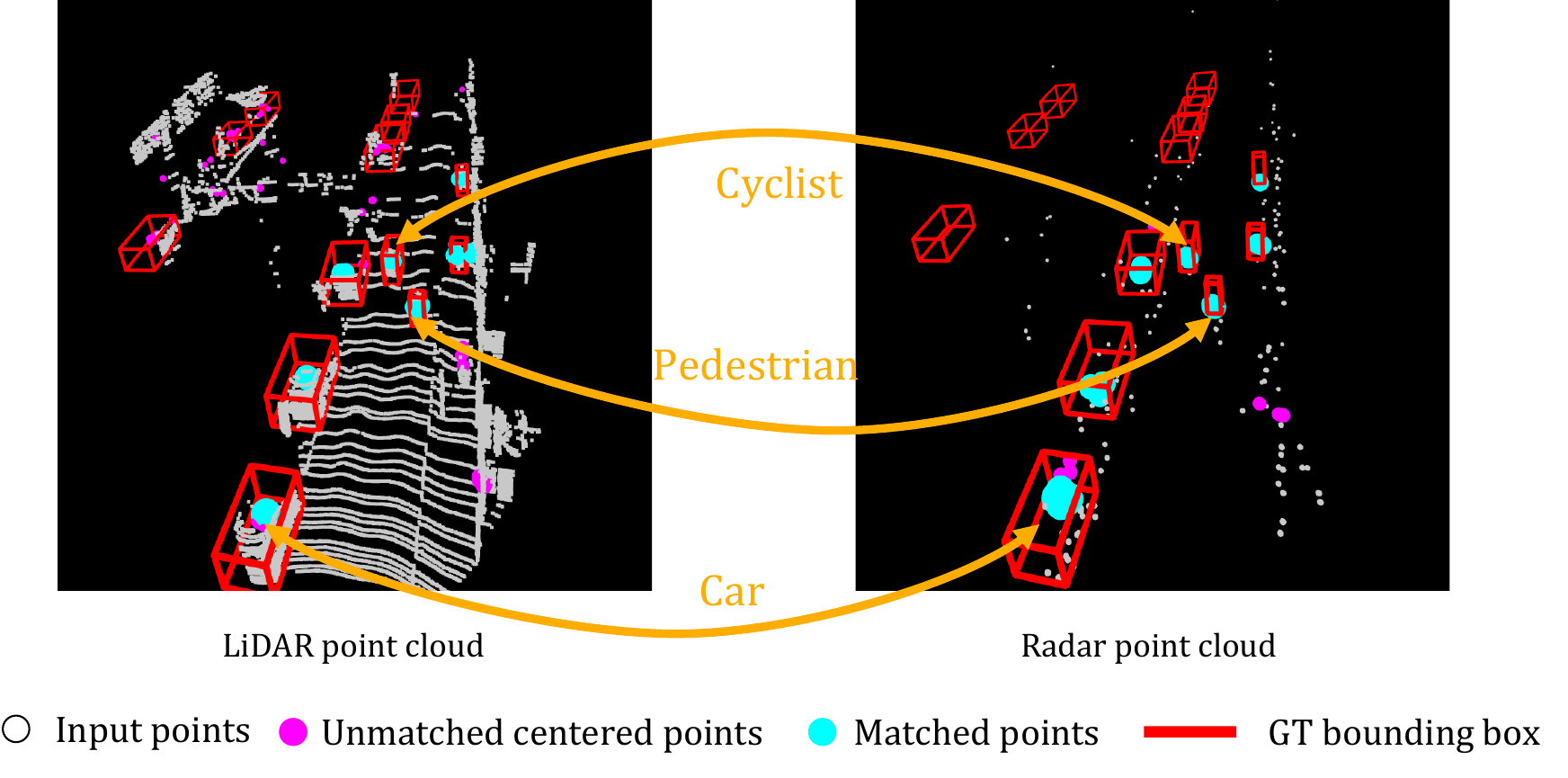}
        \caption{Visualization of the training process for selective matching. }
        \label{fig:training_vis}
    \end{subfigure}
    \caption{Here is the illustration and visualization of the selective matching during training. In Fig.~\ref{fig:training_vis}, we can see that all matched points (\textcolor{cyan}{cyan points}) are positioned near the bounding box center for co-visible objects, which demonstrates the effectiveness of the instance feature aggregation module. For wrongly sampled centered points, they will be moved to random positions and will not interfere with the training process (\textcolor{magenta}{magenta points}).}
    \vspace{-1.5em}
\end{figure}

Using the projected features, we must match pairs between modalities for training, as shown by the purple block in Fig.\ref{fig:overview}. As we note that noisy and incorrectly classified points tend to cluster at random spots, our approach involves a cross-modal Nearest Neighbor (1-NN) search within a specific radius to pinpoint the right matches, as shown in Fig.~\ref{fig:selective_matching}. Notice that only foreground points successfully shifted towards object centers can reach neighbor points from the other modality, as they are close in space. This proximity not only aids in efficient cross-modal representation learning but also reduces distractions from noisy data points.
The cross-modal feature matching loss is as follows: 
\begin{equation}
\centering
  \mathcal{L}_{FM} = \frac{1}{N_p} \sum_{i=0}^{N_f} \sum_{j=0}^{\hat{N}_f} ||h_i - \hat{h}_i||_{2} \cdot \mathbf{PM}_{ij}
\label{eqn:feature_matching}
\end{equation}
where $\mathbf{PM} \in \mathbb{R}^{N_f \times \hat{N}_f}$ is a binary matrix, and $\mathbf{PM}_{ij} = 1$ when $c_i$ and $\hat{c}_j$ are a selected matching pair, otherwise $\mathbf{PM}_{ij} = 0$. $N_p$ denotes the total number of matched pairs. 

% Note that for the cross-modal feature, there is a risk of falling into a trivial solution if the projected features from both modalities become zero vectors. We thus use an extra detection as a curator, which will be discussed in Sec.~\ref{sub:det_head}.

% \begin{figure}[t!]
%     \centering
%     \includegraphics[width=0.85\linewidth]{figure/selective_matching.pdf}
%     \caption{Illustration of selective matching. Points with lighter colors on the right are ignored in the match-searching process. 
%     }
%     \label{fig:selective_matching}
%     % \vspace{-1.5em}
% \end{figure}

\subsection{Detection Head}
\label{sub:det_head}
Following \cite{yang20203dssd, zhang2022not}, we encode bounding boxes as multidimensional vectors comprising locations, scale, and orientation. The detection head, shown as the \textcolor{det_head}{blue block} in Fig.~\ref{fig:overview}, has branches for confidence prediction and box refinement. The detection head's loss function is:
\begin{equation}
    \mathcal{L}_{Det} = \mathcal{L}_{ref} + \mathcal{L}_{cls}
\end{equation}
To avoid trival solution for feature matching in Eq.~\ref{eqn:feature_matching}, we use an extra detection head specifically for the shared space features, called shared detection head (the \textcolor{shared_head}{orange block} in Fig.~\ref{fig:overview}). This loss term is denoted as $\mathcal{L}_{S-Det}$. 

\subsection{Training and Inference}
We adopt a two-step training strategy for stability. Unlike other methods \cite{saputra2020deeptio, xu2017learning}, our feature projection branch is co-optimized with the backbone in the second phase. In the first step, we train a basic detection network (shown by the \textcolor{training_step_1}{green line} in Fig.\ref{fig:overview}) with loss:
\begin{equation}
    \mathcal{L}_{s1} =  \mathcal{L}_{Ctr} +  \mathcal{L}_{O-Reg} + \mathcal{L}_{Det}
    \label{eq:loss_stage_1}
\end{equation}
In the second step, parameters from the first phase initialize the backbone and feature aggregation module. While the auxiliary backbone remains static, the primary one undergoes joint refinement with the projection module. A new detection head addresses the augmented feature space (represented by the \textcolor{orange}{orange line} in Fig.\ref{fig:overview}). The associated loss is:
\begin{equation}
\mathcal{L}_{s2} = \mathcal{L}_{s1} + \lambda_1 \mathcal{L}_{FM} + \lambda_2 \mathcal{L}_{S-Det}
\label{eqn:s2}
\end{equation}
When it comes to inference, only the primary modality data is applied, as shown by the \textcolor{blue}{blue line} in Fig.~\ref{fig:overview}.

\section{Evaluation}
% We next move to the evaluation and performance comparison. In particular, we leverage the recent availability of a co-located LiDAR and 4D radar dataset \cite{vod} and comprehensively evaluate the effectiveness of our method on both sensor modalities. \emph{Note that our code and models will be made public based upon acceptance. }
% \jn{following structure in PVRCNN}
\subsection{Experimental Setup} \noindent \textbf{Dataset}. The View-of-Delft (VoD) dataset \cite{vod} offers calibrated, synchronized data from LiDAR, RGB camera, and 4D radar for 3D object detection, comprising 8693 frames captured in Delft's urban environment. Unique to VoD is its inclusion of vulnerable road users like pedestrians and cyclists. Notably, VoD alone provides public access to simultaneous 4D radar recordings and LiDAR data and is thus selected for evaluation. While datasets such as nuScenes \cite{caesar2020nuscenes} have both radar and LiDAR data, their radars only offer 2D spatial measurements, limiting them for 3D object detection.

\noindent \textbf{Implementation details}. We follow the design of the backbone network in \cite{zhang2022not}, which comprises several \emph{SA layers} with center-aware sampling to remove noisy points. Next, the \emph{Vote Layer} predicts an offset for each point to concentrate them on the corresponding object center for spatial alignment in centroid generation. After that, another \emph{SA layer} is employed to aggregate instance-level features. Once we obtain point clusters for different objects, we project instance-level features to a shared subspace using a 4-layer MLP. These projected features are later concatenated back to the instance-level features for bounding box classification and regression. We use $\lambda_1=\frac{1}{3}$ and $\lambda_2 = \frac{2}{3}$ in Eqn.~\ref{eqn:s2} for all our experiments. Only the primary modal data will be used during inference. The best LiDAR and radar detection models are selected based on the best validation result for their respective detection tasks. % Results for baseline models are retrained on VoD based on official implementation. 

% \subsection{Implementation Details}

%Our pipeline is trained in two stages. In the first stage, we train the detection network for the primary and auxiliary modalities without the feature projection module and select the best performance model based on the validation set. In the second stage, the pre-trained backbone weights are used for initialization and fine-tuning the whole pipeline with the projection branch and a new detection head. The best LiDAR and radar detection models are selected based on the best validation result for their respective detection tasks. Only the primary modal data will be used during inference. More detailed implementation details can be found in the supplementary. % \chris{@JN, In case reviewers ask us in rebuttal, I suggest we make clear here that, the best lidar and radar detection models are selected based on the best validation result for their respective detection tasks.}
\begin{table}[t!]
    \centering
    % \vspace{2pt}
    \resizebox{0.9\columnwidth}{!}{%
    \SetTblrInner{rowsep=1pt,colsep=2pt}
    \begin{tblr}{|l|c|c|c|c|c|}
    \hline
     \SetCell[r=2]{c}Method &\SetCell[r=2]{c} Type & Car & Pedestrian & Cyclist &  \SetCell[r=2]{c} mAP \\
      &&(IoU = 0.5) & (IoU = 0.25) & (IoU = 0.25) &  \\
     \hline
    PointPillars$^\dagger$\cite{pointpillar} & \SetCell[r=3]{c} V& 35.90&	34.90&	43.10&	38.00\\
    SECOND\cite{yan2018second} && 35.07 & 25.47 & 33.82 &31.45\\
    \underline{CenterPoint}\cite{yin2021center} && 32.32 & 17.37 & 40.25 & 29.98\\
    \hline
     \underline{PV-RCNN}\cite{shi2020pv} &PV & \textbf{38.30}&	30.79&	46.58&	38.56\\
    \hline
    \underline{PointRCNN}\cite{pointrcnn}&\SetCell[r=4]{c} P & 15.99&	34.01&	26.50&	25.50\\
    3DSSD\cite{yang20203dssd} &&  23.86 & 9.09 & 32.20 & 21.71 \\
    IA-SSD\cite{zhang2022not} && 31.33 & 23.61 & 49.58 &34.84\\
    % \hline
    \textbf{Ours} && 32.32 & \textbf{42.49} & \textbf{50.49} & \textbf{41.77}\\ 
    \hline
    \end{tblr}
    }
    \caption{Test results for \textbf{radar object detection} on VoD dataset. Note that results for PointPillars with symbol $^\dagger$ are reported in \cite{vod}. The \emph{`Type'} column denotes the data representation used in the method: \emph{`V'} denotes voxel, \emph{`P'} denotes point, \emph{`PV'} denotes point-voxel. Methods \underline{underlined} are all two-stage detectors.}
    \label{tab:test_result_radar}
    \vspace{-0.5em}
\end{table}

\begin{table}[t!]
    \centering
    \resizebox{0.9\columnwidth}{!}{%
     \SetTblrInner{rowsep=1pt,colsep=2pt}
    \begin{tblr}{|l|c|c|c|c|c|}
    \hline
     \SetCell[r=2]{c}Method&\SetCell[r=2]{c} Type& Car & Pedestrian & Cyclist &  \SetCell[r=2]{c} mAP \\
      &&(IoU = 0.5) & (IoU = 0.25) & (IoU = 0.25) &  \\
     \hline
    PointPillars$^\dagger$\cite{pointpillar} &\SetCell[r=3]{c}V & 75.60 & 55.10 & 55.40 &62.10\\
    SECOND\cite{yan2018second} && 77.69&	59.95&	65.50&	67.71\\
    \underline{CenterPoint}\cite{yin2021center} && 68.29&	66.90&	64.42&	66.54\\
    \hline
    \underline{PV-RCNN}\cite{shi2020pv} &PV & 75.16&	65.24&	66.09&	68.83\\
    \hline
     \underline{PointRCNN}\cite{pointrcnn} &\SetCell[r=4]{c} P & 61.51&	\textbf{67.36}&	67.03&	65.30\\
    3DSSD\cite{yang20203dssd} &&  77.34 & 12.64 & 37.68 &42.55\\
    IA-SSD\cite{zhang2022not} &&  77.29 & 32.18 & 57.11 &55.53\\
    % \hline
    \textbf{Ours} & & \textbf{79.74} & 60.58 & \textbf{68.52} & \textbf{69.62}\\ 
    \hline
    \end{tblr}
    }
    \caption{Test results for \textbf{LiDAR object detection} on VoD dataset. Note that results for PointPillars with symbol $^\dagger$ are reported in \cite{vod}. The \emph{`Type'} column denotes the data representation used in the method: \emph{`V'} denotes voxel, \emph{`P'} denotes point, and \emph{`PV'} denotes point-voxel. Methods \underline{underlined} are all two-stage detectors.}
    \label{tab:test_result_lidar}
    \vspace{-1em}
\end{table}

\subsection{Overall Results}

\begin{figure*}[t!]
  \centering
  \begin{subfigure}[h]{0.3\linewidth}
    \includegraphics[width=\linewidth]{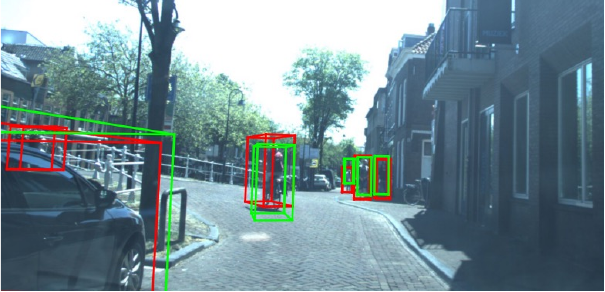}
  \end{subfigure}
  \begin{subfigure}[h]{0.3\linewidth}
    \includegraphics[width=\linewidth]{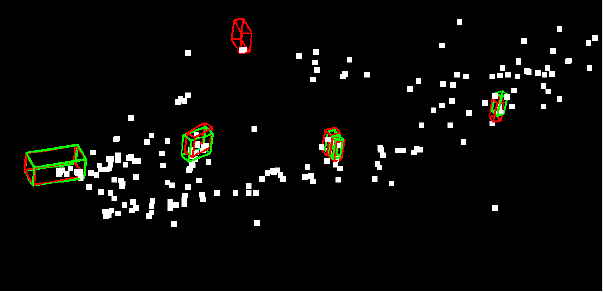}
  \end{subfigure}
  \begin{subfigure}[h]{0.3\linewidth}
    \includegraphics[width=\linewidth]{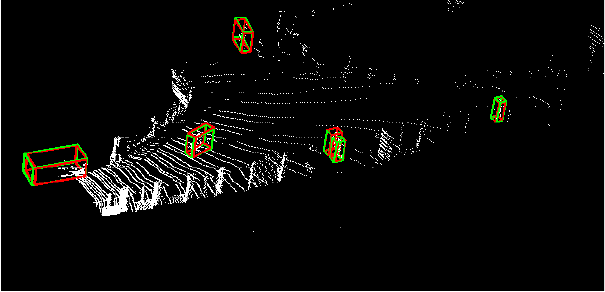}
  \end{subfigure}
  \begin{subfigure}[h]{0.3\linewidth}
    \includegraphics[width=\linewidth]{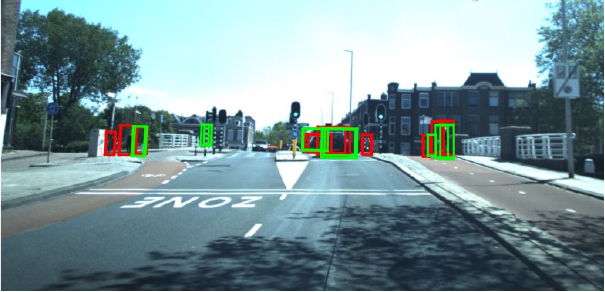}
    \caption{Radar detection in perspective}
  \end{subfigure}
  \begin{subfigure}[h]{0.3\linewidth}
    \includegraphics[width=\linewidth]{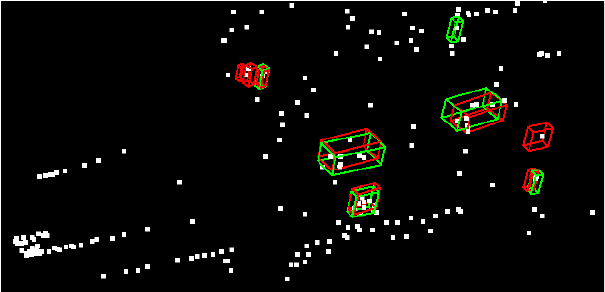}
    \caption{Radar detection}
  \end{subfigure}
  \begin{subfigure}[h]{0.3\linewidth}
    \includegraphics[width=\linewidth]{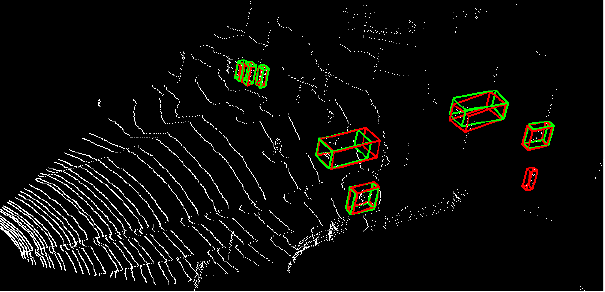}
    \caption{LiDAR detection}
  \end{subfigure}
  \caption{Qualitative result of our method. Bounding boxes for GTs are denoted in \textcolor{red}{red}, and the predictions are denoted in \textcolor{green}{green}. The left images and the middle figures are the radar detection results. Notice that the RGB images here are \textbf{only for visualization purposes but not used in model training/inference}. The right figures visualize the LiDAR point clouds and the prediction results.}
  \label{fig:qualitative_main}
  \vspace{-1em}
\end{figure*}

Tab.~\ref{tab:test_result_radar} and Tab.~\ref{tab:test_result_lidar} show that our proposed framework gives rise to the best overall performance for both modalities, achieving $41.77$ mAP and $69.62$ mAP for radar and LiDAR object detection, respectively. Fig.~\ref{fig:qualitative_main} illustrates the qualitative results of our method.

\noindent \textbf{Radar Object Detection}. Despite the challenges of sparse and noisy radar point clouds, our detection method significantly outperforms SOTA techniques. PV-RCNN \cite{shi2020pv} excels in car detection due to reduced size ambiguity in voxel representation on sparse radar clouds. Yet, our method excels in detecting smaller objects like cyclists and pedestrians. Specifically, our model scores 42.49 mAP for pedestrians, outpacing the PV-RCNN by roughly 11.7\%. This suggests that better representation for small objects with sparse clouds can be gleaned from cross-modal data, especially LiDAR. With our projection branch detailed in Sec.~\ref{sec:feature_alignment}, we encourage the radar detection network to emulate LiDAR's instance representation, and our results affirm its efficacy.

\noindent \textbf{LiDAR Object Detection}. In LiDAR detection, our technique leads in car and cyclist categories, surpassing the runner-up by 2.6\% and 2.2\% mAP respectively. This boost stems from the Radar Cross Section (RCS) features' semantic cues, distinguishing between metal and human skin\cite{kamel2017rcs,lee2016radar}. While two-stage detectors, such as \cite{yin2021center, shi2020pv, pointrcnn}, often excel in pedestrian detection through second-stage ROI refinement, they come with larger memory and slower speeds. 

\subsection{Ablation Study}

To be consistent with LiDAR detection work, the more demanding 3D IoU in KITTI \cite{kitti} with 0.7, 0.5, and 0.5 is used for LiDAR evaluation and 0.5, 0.25, 0.25 for radar evaluation on the VoD \cite{vod} \emph{validation} set. 

\noindent \textbf{Effect of auxiliary-modal attributes}. Through experiments with different auxiliary modal data inputs, we assess their influence on radar object detection using cross-modal supervision. Tab.~\ref{tab:ablation_radar_with_LiDAR} shows that radar detection improves when incorporating the LiDAR point cloud's geometric information ($x$, $y$, $z$). The LiDAR intensity attribute has a slight impact, given radar's stable semantic cues from RCS measurements. The Car category's modest performance gain is due to around 25\% of cars lacking radar points, hindering detection. Moreover, cars' larger size versus cyclists and pedestrians complicates bounding box estimation with few points.

% \begin{table}[t!]
% \centering
% \resizebox{0.9\columnwidth}{!}{%
% \SetTblrInner{rowsep=1pt,colsep=2pt}
% \begin{tabular}{l|cccc}
% \hline
% \multicolumn{1}{c|}{\begin{tabular}[c]{@{}c@{}}Auxiliary radar \\ point attributes\end{tabular}} &
%   \begin{tabular}[c]{@{}c@{}}Car \\ (IoU=0.7)\end{tabular} &
%   \begin{tabular}[c]{@{}c@{}}Pedestrian\\ (IoU=0.5)\end{tabular} &
%   \begin{tabular}[c]{@{}c@{}}Cyclist\\ (IoU=0.5)\end{tabular} &
%   \multicolumn{1}{l}{mAP} \\ \hline
% + none (baseline)      & 48.86 & 48.91          & 67.88          & 55.22         \\
% + $x,y,z$              & 58.98          & 48.81          & 78.04          & 61.94         \\
% + $x,y,z,RCS$          & 58.82          & \textbf{56.29} & \textbf{78.17} & \textbf{64.42} \\
% + $x,y,z,v$     & \textbf{59.27}          & 50.14          & 77.91          & 62.44          \\
% + $x,y,z,RCS,v$ & 59.18          & 55.30          & 77.56          & 64.01          \\ \hline
% \end{tabular}}
% \caption{LiDAR detection results in \emph{validation set} supervised by different radar attributes during training. $x, y, z$ for point position, $RCS$ for radar cross section, $v$ for velocity.}
% \label{tab:ablation_LiDAR_with_radar}
% \vspace{-0.5em}
% \end{table}

\begin{table}[t!]
\centering
\resizebox{0.9\columnwidth}{!}{%
\SetTblrInner{rowsep=1pt,colsep=2pt}
\begin{tblr}{l|cccc}
\hline
\SetCell[r=1]{c}Auxiliary LiDAR & Car & Pedestrian & Cyclist & \SetCell[r=2]{c} mAP\\
\SetCell[r=1]{c}point attributes & (IoU=0.5) & (IoU=0.25) & (IoU=0.25)&  \\
\hline
+ none (baseline)      & 31.24 & 32.50          & 60.69          & 41.48         \\
+ $x,y,z$              & 31.32          & 40.04          & 67.78          & 46.38         \\
+ $x,y,z,I$          & \textbf{32.20}          & \textbf{40.42} & \textbf{68.67} & \textbf{47.03} \\
\hline
\end{tblr}}
\caption{Radar detection results in \emph{validation set} supervised by different LiDAR attributes during training. $x, y, z$ for point position, $I$ for intensity.}
\label{tab:ablation_radar_with_LiDAR}
% \vspace{-1em}
\end{table}

\noindent   \textbf{Effect of joint optimization}. We study the effectiveness of our joint optimization strategy.

% two major points: 1. our methods yields a more robust backbone network. 2. the projection branch successfully encodes imagery features from the other modality

\noindent   \textit{(a). Setup}. We use the base network trained without cross-modal supervision and the projection branch as the ablation baseline. Later, we remove the projection branch of our framework shown in Fig.~\ref{fig:overview}, freeze the backbone weights, and train a new detection head for performance comparison. 

\begin{figure}[t!]
  \centering
  \begin{subfigure}[b]{0.45\linewidth}
    \includegraphics[width=\linewidth]{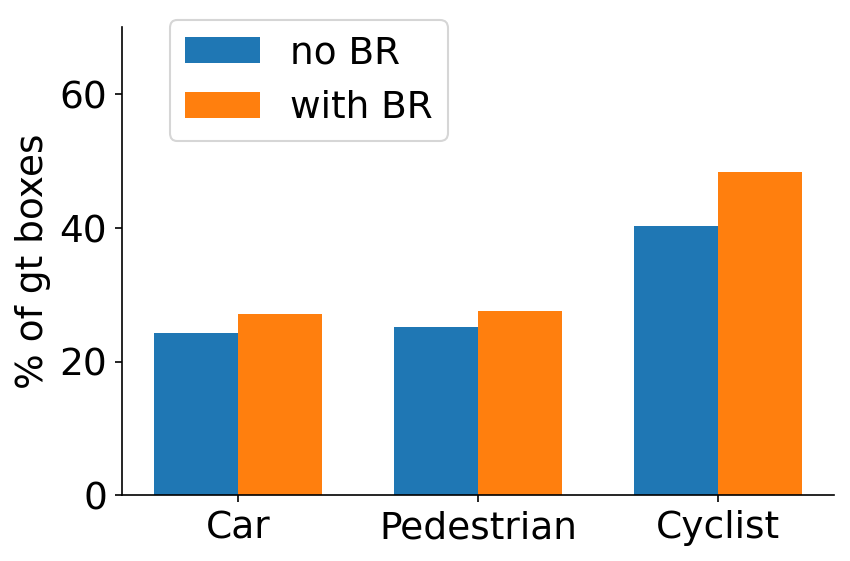}
    \caption{Radar backbone statistics}
  \end{subfigure}
  \begin{subfigure}[b]{0.45\linewidth}
    \includegraphics[width=\linewidth]{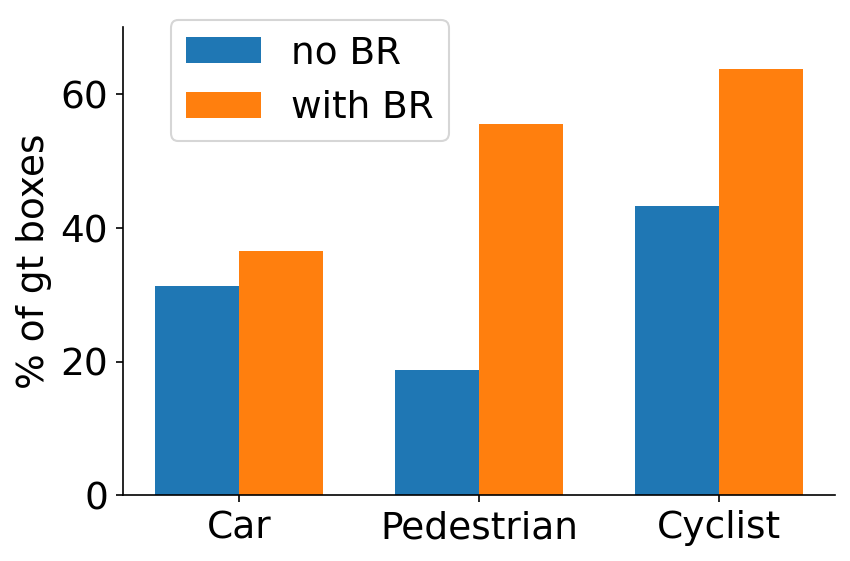}
    \caption{LiDAR backbone statistics}
  \end{subfigure}
  \caption{Percentage of instance containing 5-20 points in the backbone output. Left for \textbf{radar}, right for \textbf{LiDAR}, The \textbf{BR} denotes backbone refinement.}
  \label{fig:backbone_refinement}
  % \vspace{-1em}
\end{figure}

\noindent   \textit{(b). Results}.  Looking at the first two rows in Tab.~\ref{tab:Ablation_projection}, mAP increases in both modalities, even without the projection branch. Significant improvements are seen with the Cyclist on radar and Pedestrian on LiDAR. For LiDAR, fewer car matches might cause a minor decline with only \textbf{BR} (Backbone Refinement), but other categories enhance. To understand why only refined backbone improves performance, we collected backbone output statistics. We can see from Fig.~\ref{fig:backbone_refinement} that the percentage of instances containing more than five points gets improved. We propose that cross-modal representation learning enhances the backbone network's feature extraction. These robust features are partially propagated to close points via max pooling, retaining more significant instance points in the final sampling layer, thus creating a robust instance feature representation for improved detection. % More detailed statistics are in the supplementary.
Furthermore, the last row in each modality shows that the entire network with the projection branch leads to the best performance for all three categories, indicating its effectiveness and that augmented cross-modal features can be encoded in our model for better detection inference. 

\begin{table}[t!]
    \centering
    \resizebox{0.9\columnwidth}{!}{%
    \SetTblrInner{rowsep=1pt,colsep=2pt}
    \begin{tblr}{c|cc|cccc}
        \hline
         \SetCell[r=2]{c}Modality & \SetCell[r=2]{c}\textbf{BR} & \SetCell[r=2]{c}\textbf{P} & Car& Pedestrian & Cyclist & \SetCell[r=2]{c}mAP  \\
         & & & (IoU=0.5) & (IoU = 0.25) & (IoU = 0.25) &\\
         \hline
         \SetCell[r=3]{c}radar & & & 31.24 & 32.50 & 60.69 & 41.48 \\
        %  \hline
         & \SetCell[r=1]{c}$\checkmark$  &  & 31.51& 33.48& 67.19 & 43.94\\
        %  & & &  (\textbf{+0.27}) & (\textbf{+0.98)} &(\textbf{+6.50}) & (\textbf{+2.46})\\
        %  \hline
          & \SetCell[r=1]{c}$\checkmark$  & \SetCell[r=1]{c}$\checkmark$ & \textbf{32.20} & \textbf{40.42}  & \textbf{68.67}         & \textbf{47.03} \\
        %   & & &(\textbf{+0.78}) &(\textbf{+6.94}) & (\textbf{+7.98})  & (\textbf{+5.55})\\
          \hline
          \hline
        %   &  &  & Car& Pedestrian & Cyclist & \SetCell[r=2]{c}mAP  \\
         & & & (IoU=0.7) & (IoU = 0.5) & (IoU = 0.5) & mAP\\
         \hline
         \SetCell[r=3]{c}LiDAR&  &       & 57.94&	29.40&	68.26&	51.87 \\
        %  \hline
          & \SetCell[r=1]{c}$\checkmark$&  & 55.72    & 48.64 & 76.22  & 60.20\\			
        %   & & & (\textbf{-2.21}) & (\textbf{+19.24}) & (\textbf{+7.97}) & (\textbf{+8.33})\\
        %  \hline
         &\SetCell[r=1]{c}$\checkmark$ & \SetCell[r=1]{c}$\checkmark$    & \textbf{58.82}	&\textbf{56.29}	&\textbf{78.17}	&\textbf{64.42}   \\	
        %  & & & (\textbf{+0.88}) & (\textbf{+26.88}) & (\textbf{+9.91}) & (\textbf{+12.56})\\
          \hline
    \end{tblr}
    }
    \caption{Evaluation on the validation set. \textbf{BR}: network with backbone refinement which optimizes the backbone in the second step training with cross-modal supervision, \textbf{P}: network with projection branch.}
    \label{tab:Ablation_projection}
    \vspace{-1.5em}
\end{table}

% \subsubsection{Effect of two-level alignment}
\noindent   \textbf{Effect of two-level alignment}. We next study the LiDAR object detection results to better understand the influence of our alignment strategies introduced in Sec.~\ref{sec:spatial_alignment} and Sec.~\ref{sec:feature_alignment}. Tab.~\ref{tab:ablation_alignment} shows that spatial alignment in centroid generation is the key to cross-modal learning. When removed, the discrepancy between modalities results in mismatching pairs. Misleading supervision signals caused by mismatched pairs degenerate the network performance, especially for large objects like Cars. 
When the spatial alignment in centroid generation is added, the model establishes correct cross-modal instance matches and improves mAP, as shown in the third row of Tab.~\ref{tab:ablation_alignment}. We attribute the slight performance drops on cars to fewer matched instances in this category.
As expected, the feature alignment will only function and positively contribute to the model when the cross-modal points are spatially aligned first. The last row shows that combining two alignment operations yields the best performance. Similar conclusions are observed for radar detection, and we omit its discussion to avoid repetition.

\begin{table}[t!]
    \centering
    \resizebox{0.9\columnwidth}{!}{
    \SetTblrInner{rowsep=1pt,colsep=2pt}
    \begin{tblr}{c|cc|cccc}
    \hline
    \SetCell[r=2]{c}{Modality} & \SetCell[r=2]{c} Spatial & \SetCell[r=2]{c} Feature & Car & Pedestrian & Cyclist & \SetCell[r=2]{c}mAP\\
    % \hline
     & & & (IoU=0.7) & (IoU=0.5) & (IoU=0.5) & \\
     \hline
     \SetCell[r=4]{c}LiDAR &  &  &57.94&	29.40&	68.26&	51.87 \\
    %  \hline
     &  & $\checkmark$ &28.88&	30.22&	64.33&	41.15 \\
    %  \hline
     & $\checkmark$ & & 57.29&	48.66&	77.70&	61.22\\
    %  \hline
     &$\checkmark$ & $\checkmark$& \textbf{58.82}&	\textbf{56.29}&	\textbf{78.17}&	\textbf{64.42}\\
     \hline
     
    \end{tblr}
    }
    \caption{Ablation on the effect of spatial alignment in centroid generation and feature level alignment. Notice that we use the basic model trained without any cross-modal information in Tab.~\ref{tab:Ablation_projection} as the baseline performance.}
    \label{tab:ablation_alignment}
    \vspace{-0.5em}
\end{table}

\subsection{Runtime Efficiency}
\label{sec:runtim_efficiency}
\begin{table}[t!]
    \centering
    \resizebox{0.8\columnwidth}{!}{
    \SetTblrInner{rowsep=1pt,colsep=2pt}
    \begin{tblr}{c|c|c|c|c}
    \hline
          Type & Method & Memory & Parallel & Speed \\
         \hline
          \SetCell[r=3]{c}Voxel-based & PointPillars\cite{pointpillar}& 354MB & 69 & 123 \\
         \hline
          & SECOND\cite{yan2018second}& 710MB & 34 & 34 \\
          \hline
          & CenterPoint\cite{yin2021center}& 204MB & 119 & 66 \\
         \hline
         Point-Voxel & PV-RCNN\cite{shi2020pv}& 1223MB & 17 & 13 \\
         \hline
         \SetCell[r=4]{c}Point-based& pointRCNN\cite{pointrcnn} &  560MB & 43 & 14  \\
         \hline
          & 3DSSD\cite{yang20203dssd} & 502MB & 48 & 20 \\
         \hline
          & IA-SSD\cite{zhang2022not} & 120MB & 202 & 23 \\
         \hline
          & \textbf{Ours} & \textbf{133MB} & \textbf{183} & \textbf{23} \\
         \hline
    \end{tblr}
    }
    \caption{ Comparing memory usage and runtime efficiency with LiDAR input. Memory for each method is measured using the same 16384 points per scan. 'Parallel' refers to the maximum batch size for one RTX 3090. Speed is gauged with single scan input and reported in frames per second. Our method's results are highlighted in \textbf{bold}.} 
    \label{tab:efficiency}
    \vspace{-1.5em}
\end{table}

We evaluate the memory consumption and inference speed compared with SOTA methods with LiDAR as the primary modal input. As shown in Tab.~\ref{tab:efficiency}, our method has the second lowest memory footprint among all methods and the fastest inference speed in the point-based methods. Together with Tab.~\ref{tab:test_result_lidar} and Tab.~\ref{tab:test_result_radar}, our framework has the best balance between efficiency and accuracy.

\section{Conclusions and Future Work}

This work introduced a novel framework to improve the robustness of single-modal 3D object detection via cross-modal supervision. Our method is able to effectively exploit the side information from auxiliary modality data for a better-informed backbone network and a robust hallucination branch. Bespoken spatial and domain alignment strategies are also proposed to address the fundamental discrepancy across modalities and showed a significant performance improvement. Experimental results on VoD\cite{vod} demonstrate that our method outperforms SOTA methods in object detection for both radar and LiDAR while maintaining a competitive memory footprint and runtime efficiency.

%%%%%%%%%%%%%%%%%%%%%%%%%%%%%%%%%%%%%%%%%%%%%%%%%%%%%%%%%%%%%%%%%%%%%%%%%%%%%%%%
\vspace{1em}
\bibliographystyle{IEEEtran}
\bibliography{reference}

\end{document}